\documentclass{article}
\usepackage{ijcai19}

\usepackage{times}
\usepackage{soul}
\usepackage{url}
\usepackage[hidelinks]{hyperref}
\usepackage[utf8]{inputenc}
\usepackage[small]{caption}
\usepackage{graphicx}
\usepackage{amsmath}
\usepackage{booktabs}
\usepackage{algorithm}
\usepackage{algorithmic}
\usepackage{algorithm,algorithmic}

\usepackage{amssymb}

\usepackage{comment}

\usepackage{amsthm}

\usepackage{subfig}
\usepackage{caption}

\usepackage{mathtools}

\usepackage{bm}

\usepackage{array}
\usepackage{booktabs}
\usepackage{multirow}

\newtheorem*{defi}{Definition}

\title{Curriculum Learning for Cumulative Return Maximization}

\author{
Francesco Foglino
\and
Christiano Coletto Christakou\and
Ricardo Luna Gutierrez\and
Matteo Leonetti
\affiliations
School of Computing, University of Leeds, United Kingdom\\
\emails
\{scff,mm18ccc,scrlg,m.leonetti\}@leeds.ac.uk
}

\begin{document}

\maketitle

\begin{abstract}
Curriculum learning has been successfully used in reinforcement learning to accelerate the learning process, through knowledge transfer between tasks of increasing complexity. Critical tasks, in which suboptimal exploratory actions must be minimized, can benefit from curriculum learning, and its ability to shape exploration through transfer. We propose a task sequencing algorithm maximizing the cumulative return, that is, the return obtained by the agent across all the learning episodes. By maximizing the cumulative return, the agent not only aims at achieving high rewards as fast as possible, but also at doing so while limiting suboptimal actions. We experimentally compare our task sequencing algorithm to several popular metaheuristic algorithms for combinatorial optimization, and show that it achieves significantly better performance on the problem of cumulative return maximization. Furthermore, we validate our algorithm on a critical task, optimizing a home controller for a micro energy grid. 
\end{abstract}

\section{Introduction}
Curriculum learning (CL) has gained popularity in reinforcement learning as a means to guide exploration in complex tasks~\cite{shao2018micromancl,doom2017}. The agent is led to learn, in simple tasks, knowledge that can be successfully generalized and exploited in larger tasks.  A central aspect of curriculum learning is task sequencing, since the order in which the tasks are executed is a major factor in the quality of a curriculum. The goal of automatic curriculum generation, so far, has been to reach the optimal policy faster, with sequencing algorithms minimizing a transfer learning metric called \emph{time-to-threshold}~\cite{tl4rl}. Time-to-threshold measures the total training time, along the whole curriculum, to achieve a given performance threshold (in terms of cumulative reward). Therefore, having an estimate of the quality of the optimal policy, it is possible to obtain a curriculum which reaches the optimal policy faster than learning from scratch. However, in this setting, while the time to reach the optimal policy is minimized, the behavior of the agent \emph{during} learning is not taken into account.

We consider the novel case, for CL, in which an agent needs to learn online in a critical task, in which exploration is costly, and therefore the number of suboptimal actions during exploration should be minimized. An example of this setting is control for smart-grid agents, which while learning consume real energy from the grid. For this class of problems, it is desirable not only to learn the optimal policy quickly, but also to reach it with the smallest possible amount of suboptimal exploration. 
\begin{figure}[htb]
	\includegraphics[width=\linewidth]{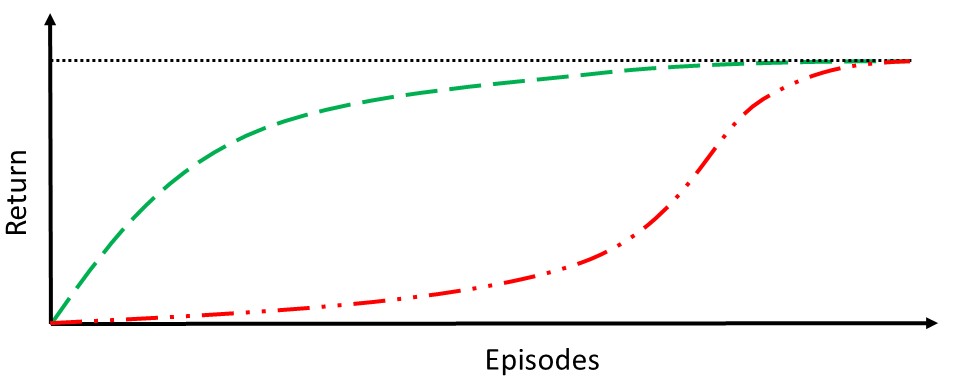}
	\caption{Two learning curves with the same time-to-threshold. The curve above is preferable in critical tasks.
	}
	\label{Fig:ttt_illustration}
\end{figure}
By means of illustration, consider the graph in Figure \ref{Fig:ttt_illustration}. The two learning curves have the same time-to-threshold, but one achieves a higher return than the other in every episode. Such a learning behavior is preferable, and is the objective of our optimization.

In this paper, we introduce a novel use of curriculum learning, in order to provide the optimal initial knowledge resulting in the minimum amount of exploration in critical tasks. We propose to adopt the \emph{cumulative return}, rather than time-to-threshold, as the objective function for task sequencing. Furthermore, we introduce a novel heuristic search algorithm tailored for the resulting sequencing problem, and we experimentally show that it outperforms three popular metaheuristic algorithms, on several tasks. Lastly, we validate our method on a real critical domain: the optimization of a home controller for a micro-grid environment. We show that through curriculum learning, the agent can save a considerable amount of energy during exploration.

\section{Related Work}

Curriculum Learning  in reinforcement learning is an increasingly popular
field, with successful examples of applications in first-person
shooter games \cite{doom2017,wu2018}, real-time strategy games \cite{shao2018micromancl},
and real-world robotics applications \cite{pmlr-v80-riedmiller18a}.

Curricula can be considered within a particular environment, where the curriculum is obtained, for instance, by sequencing the initial and goal states~\cite{asada1996purposive,sukhbaatar2018,florensa2018goalgen}, or by defining additional auxiliary tasks~\cite{pmlr-v80-riedmiller18a}. Curricula can also be generated at the task level, 
where the tasks to be scheduled may have different environment dynamics, and the agent learns in one environment, before moving on to the next. Our sequencing approach is designed for the latter case.

In task-level CL, the knowledge transfer between different environments plays a crucial role. Similar problems have been considered in multi-task reinforcement learning \cite{wilson2007multi}, and lifelong learning \cite{ruvolo2013ella}, where the agent attempts to maximize its performance over the entire set of tasks, which may not all be related to one another.  Conversely, in curriculum learning, the intermediate tasks are generated specifically to be part of the curriculum, and the curriculum is optimized for a pre-specified set of final tasks.

The automatic generation of curricula~\cite{da2019survey} has been divided into two sub-problems: task generation~\cite{leonetti16,daSilva2018}, that is the problem of creating a set of tasks such that transferring from them is most likely beneficial for the final task; and task sequencing~\cite{leonetti17,narvekarIJCAI17,daSilva2018,foglino2019optimization}, whereby previously generated tasks are optimally selected and ordered. Current methods for task sequencing attempt to determine the optimal order of tasks
either with~\cite{narvekarIJCAI17,baranes2013active} or without
\cite{leonetti17,daSilva2018} executing the tasks. All task sequencing methods mentioned above are heuristic algorithms tailored to the minimization of time-to-threshold. In this paper, we propose to maximize the cumulative return instead, and propose a novel heuristic algorithm for the resulting optimization problem.


\section{Background}

\subsection{Reinforcement Learning}
We model tasks as episodic Markov Decision Processes. An MDP is a tuple $\langle S,A,p,r,\gamma \rangle$, where $S$ is the set of states, $A$ is the set of actions, $p : S \times A  \times S \rightarrow [0,1]$ is the transition function, $r: S \times A \rightarrow \mathbb{R}$ is the reward function and $\gamma \in [0, 1]$ is the discount factor. 
Episodic tasks have \emph{absorbing} states, that are states that can never be left, and from which the agent only receives a reward of $0$.

For each time step $t$, the agent receives an observation of the state and takes an action according to a policy $\pi : S \times A \rightarrow [0,1]$. The aim of the agent is to find the \textit{optimal} policy $\pi^*$ that maximizes the expected discounted return $G_0 = \sum_{t=0}^{t_M} \gamma^t r(S_t, A_t)$, where $t_M$ is the maximum length of the episode. Sarsa($\lambda$) is a learning algorithm that takes advantage of an estimate of the \emph{value} function $q_\pi(s,a) = E_\pi [G_t \mid S_t = s, A_t = a]$. We represent the value function with either a linear function approximator, or a deep neural network.


Curriculum learning leverages transfer learning to transfer knowledge through the curriculum, in order to benefit a final task. Transfer takes place between pairs of tasks, referred to as the \emph{source} and the \emph{target} of the transfer. We use a transfer learning method based on value function transfer~\cite{tl4rl},  which uses the learned source q-values, representing the knowledge acquired in the source task,  to initialize the value function of the target task. 

\subsection{Combinatorial Optimization}
\label{sec:combinatorialop}
Combinatorial Optimization (CO) problems are characterized by the goal of finding the optimal configuration of a set of discrete variables. The most popular approaches in this field, called \emph{metaheuristics}, are \emph{approximate} algorithms, that do not attempt to search the solution space completely, but give up global optimality in favor of finding a good solution more quickly. Metaheuristics are applicable to a large class of optimization problems, and are the most appropriate methods for black-box combinatorial optimization, when a particular structure of the objective function (for instance, convexity) cannot be exploited. Task sequencing is one such black-box problem, therefore we selected three of the most popular metaheuristics algorithms for comparison with our search method: Tabu Search \cite{glover1989tabu}, Genetic Algorithm \cite{goldberg1989genetic}, and Ant Colony Search \cite{dorigo1991ant}. Tabu Search is a \emph{trajectory based} algorithm, which starting from a single random instance searches through the neighborhood of the current solution for an improvement. Genetic Algorithm and Ant Colony Search are \emph{population based} algorithms, that start from a set of candidate solutions, and improve them iteratively towards successive areas of interest.

\section{Problem Definition}
\label{sec:probdef}

Let $\mathcal{T}$ be a finite set of MDPs constituting the candidate \emph{intermediate} tasks. We define, in the context of this work, a curriculum as a sequence of tasks in $\mathcal{T}$ without repetitions:
\begin{defi}{\emph{[Curriculum]}} Given a set of tasks $\mathcal{T}$,
	a \emph{curriculum} over $\mathcal{T}$ of length $l$ is a sequence
	of tasks $c=\langle m_{1},m_{2},$ $\ldots,$ $m_{l}\rangle$ where each $m_{i}\in\mathcal{T}$, and $\forall i,j \in [1,l] \; i \neq j \Rightarrow m_i \neq m_j$.
\end{defi}
We assume that the agent learns each task until convergence, and that each task serves the purpose of learning one additional skill. Let $\mathcal{F}$ be a finite set of MDPs constituting the \emph{final} tasks. These are the tasks of interest, and for a curriculum to be valuable, it must provide an advantage over learning the final tasks directly. 

\subsection{Critical Task Scenario}
\label{SubSec:critical}

We target the following scenario: one or more critical tasks of interest must be learned online, by limiting suboptimal actions as much as possible. The aim of the curriculum is to provide the best possible initial knowledge so as to shape exploration in the final tasks. We assume that a simulator is available, to train the agent while generating the curriculum. We are primarily interested in optimizing the behavior of the agent for the real final tasks, and we consider the time spent generating the curriculum in simulation as a sunk cost. This setting is common to many real-world applications, where simulators are available, and real-world exploration is costly. For instance, it largely applies to robotics.

\subsection{Optimization Problem}
\label{SubSec:Optimization}

For the setting described above, we consider the most appropriate objective function to be the expected \emph{cumulative return} $\mathcal{J} : \mathcal{T} \times \mathcal{F} \rightarrow \mathbb{R}$:
\begin{equation}
\label{eq:objective}
\mathcal{J}(c,m_{f}) \coloneqq \sum_{i=1}^{N} \mathbb{E} [G_{f}^{i}], 
\end{equation}
where $G_{f}^{i}$ is the return obtained by the agent in the final task $m_{f} \in \mathcal{F}$ at episode $i$, and $N$ is the maximum number of episodes executed in the final task. Analogous objectives have been considered in the literature in the case of single-task exploration (regret~\cite{jaksch2010near}), and transfer learning (area under the curve~\cite{tl4rl}, and area ratio~\cite{lazaric2012transfer}).

Let $\mathcal{C}_{l}^{\mathcal{T}}$ be the set of all curricula over $\mathcal{T}$ of length $l$. In the rest of this paper we will drop the superscript wherever the set of candidate tasks is implicit. We define $\mathcal{C}_{\leq L} \coloneqq \bigcup_{l=1}^L C_l$ as the set of all curricula of length at most $L$.

We consider the problem of finding an optimal curriculum $c^{*}$ of a given maximum length $L$, maximizing the cumulative return over all final tasks, $\mathcal{P}(c,\mathcal{F}) = \sum_{m_f \in \mathcal{F}} \mathcal{J}(c,m_{f})$:

\begin{equation}
\label{eq:problem}
\begin{split}
\mathrm{max} \; & \mathcal{P}(c,\mathcal{F}) \\
\mathrm{s.t.}\; &  c \in \mathcal{C}_{\leq L}
\end{split}
\end{equation}

This optimization problem is entirely solved in simulation, that is, all tasks, including the final ones, are simulated tasks. Simulated final tasks are models of the expected real tasks, and having more than one prevents the curriculum from overfitting to a particular simulated task.

The return $G_{f}^{i}$ obtained by the agent in each episode for a given final task is a random variable, which depends on the dynamics of the task, the initial knowledge of the agent, and the exploration algorithms employed. The expectation cannot be computed exactly, and must be estimated from a number of trials. The resulting objective function does not have an explicit definition, therefore Problem \ref{eq:problem} is black-box, and it is in general nonsmooth, nonconvex, and even discontinuous.  Furthermore, the optimization problem is constrained to a combinatorial feasible set. 
These characteristics do not allow us to resort to methods for general Mixed-Integer NonLinear Programs, or standard Derivative-Free methods. 
The most appropriate class of optimization algorithms for this type of problem is the class of metaheuristc algorithms, introduced in Section \ref{sec:combinatorialop}.

\section{Heuristic Algorithm for Task Sequencing}

While metaheuristc algorithms are quite general and broadly applicable, it is possible to devise specific heuristic methods targeted at particular problems. In this section, we introduce Heuristic Task Sequencing for Cumulative Return (HTS-CR).

We take advantage of the following insight: the quality of a particular task sequence is strongly affected by the efficacy of knowledge transfer, and transfer between certain pairs of tasks is much more effective than others.  Therefore, our algorithm starts by considering all pairs of tasks, in order to assess which ones are good sources for which targets. It also determines which tasks are the best candidates to be \emph{head} of the curriculum, or \emph{tail}, that is, the last task before the final tasks. The method is shown in Algorithm \ref{Alg:CurrSearch}.

This first phase, expanded in Algorithm \ref{Alg:EvaluatePairs}, consists in evaluating all curricula of length $2$ (Line \ref{alg:pairs:evaluatePairs}), and sort them (Line \ref{alg:pairs:sort}), with the best curriculum first. At Line \ref{alg:pairs:headscore} and \ref{alg:pairs:tailscore} the algorithm assignes a score to each task: the better the length-2 curriculum it belongs to, the lower the score. These scores are returned by the function \texttt{EvaluatePairs}.
\begin{algorithm}[htb]
	\caption{HTS-CR}
	\label{Alg:CurrSearch}
	
	\begin{algorithmic}[1]
		\REQUIRE $\mathcal{T}$,  $\mathcal{F}$, and $L$
		\ENSURE curriculum $c^*$ and its value $v^*$
		
		\STATE $I \leftarrow 1$, $J \leftarrow 1$
		\STATE $\langle heads, tails, V \rangle \leftarrow$ \texttt{EvaluatePairs}($\mathcal{T}, \mathcal{F}$)
		
		\FOR{$r$ from 1 to  $2$($\lvert \mathcal{T} \rvert -1$)}
		\label{alg:search:rounds}
		
		\IF{($r$ mod $2$) $ = 1$}
		\STATE $I = I + 1$
		\ELSE
		\STATE $J = J + 1$
		\ENDIF
		
		\STATE $H \leftarrow$ \texttt{Best}($heads$,$I$) \COMMENT $I$ tasks with lowest score
		\label{alg:search:candidateH}
		\STATE $T \leftarrow$  \texttt{Best}($tails$,$J$) \COMMENT $J$ tasks with lowest score
		\label{alg:search:candidateT}
		
		\FOR{$l$ from 3 to $L$} 
		\label{alg:search:length}
		  \FOR{$\bar{h} \in H$ and $\bar{t} \in T$ s.t. $\bar{h} \neq \bar{t}$} 
		    
		      \STATE $\mathcal{B} \leftarrow H \cup T \setminus \{\bar{h},\bar{t}\}$
		      \STATE $B \leftarrow$ \texttt{Permutations($\mathcal{B}, l-2$)}
		      \label{alg:search:permutations}
		
		    \FOR{$b \in B$}
		    \STATE $c \leftarrow \langle \bar{h}, b, \bar{t} \rangle$
		      \IF{$c \notin V$}
		      \label{alg:search:alreadyeval}
			\STATE $v \leftarrow \mathcal{P}(c, \mathcal{F})$
			\label{alg:search:evaluate}
			\STATE $V\leftarrow P\cup\{\langle c,v \rangle\}$
		      \ENDIF
		    \ENDFOR
				  
		\ENDFOR 
		\ENDFOR
		\ENDFOR
		
		\RETURN $\langle c^*, v^*\rangle \in V$ s.t. $\forall \langle c,v \rangle \in P, v \leq v^*$ 
	\end{algorithmic}
\end{algorithm}
\begin{algorithm}[htb]
	\caption{EvaluatePairs}
	\label{Alg:EvaluatePairs}
	
	\begin{algorithmic}[1]
		\REQUIRE $\mathcal{T}$ and $\mathcal{F}$
		\ENSURE $\langle heads, tails \rangle$
		
		\STATE $D,V \leftarrow \emptyset$
		\STATE $heads = tails = [ \{m_1, 0\}, \dots,\{m_{\lvert \mathcal{T} \rvert}, 0 \} ]$ \COMMENT dictionary 
		\COMMENT from tasks to integers
		
		\STATE $D \leftarrow$ \texttt{AllPairs}($\mathcal{T}$)
		\STATE $V \leftarrow \{\langle d, v \rangle \vert d \in D \wedge v= \mathcal{P}(d,\mathcal{F})\}$  \COMMENT evaluate all pairs
		\label{alg:pairs:evaluatePairs}
		
		\STATE $V \leftarrow$ \texttt{Sort}$(V)$ \COMMENT sort wrt cumulative return, best first.
		\label{alg:pairs:sort}
		
		\FOR{$i$ from $1$ to $\lvert V \rvert$}
		
			\STATE $\langle d, v\rangle  \leftarrow V_i$ \COMMENT i-th best curriculum in V
			
			\STATE  $\langle h, t\rangle  \leftarrow d$ \COMMENT head and tail of $d$

			\STATE $heads[h] \leftarrow heads[h] + i$ 
			\label{alg:pairs:headscore}
			\STATE $tails[t] \leftarrow tails[t] + i$ 
			\label{alg:pairs:tailscore}
		\ENDFOR
		
		\RETURN $\langle heads, tails, V \rangle$
		
	\end{algorithmic}
\end{algorithm}

After this initial phase, Algorithm \ref{Alg:CurrSearch} uses the computed scores to determine the order in which curricula are evaluated. The underline intuition is the following: the most promising head and tail tasks are tried first, and shorter curricula are evaluated before longer ones. At each round $r$ (Line \ref{alg:search:rounds}), one more task is added to the set of candidate heads (Line \ref{alg:search:candidateH}) or tails (Line \ref{alg:search:candidateT}) alternatively. For each length up to the maximum length (Line \ref{alg:search:length}), and for all pairs of tasks, one from the head set $H$, $\bar{h}$, and one from the tail set $T$, $\bar{t}$, the algorithm generates all the permutations of the remaining tasks in $H \cup T$ (Line \ref{alg:search:permutations}). It then appends  $\bar{h}$ at the beginning, and $\bar{t}$ at the end, creating a full curriculum, which, if not considered before (Line \ref{alg:search:alreadyeval}) is evaluated by running the corresponding simulation, estimating the cumulative return (Line \ref{alg:search:evaluate}).

HTS-CR has no parameters other than the maximum length of the curricula to be searched, which can be exchanged for a different stopping criterion, such as a maximum budget of curricula evaluations. We intentionally left out all curricula of length $1$, since our heuristic would not have any meaningful order among them. They could be evaluated in a preliminary phase in any order.
We will show experimentally in the next section that, after the initial cost of evaluating all curricula of length $2$, the solutions found are quickly close to optimal.

\section{Heuristic Search Evaluation}

We organize the experimental evaluation into two parts. The first part, described in this section, has the aim of evaluating our heuristic search against some of the most popular optimization algorithms applicable to our problem formulation: Tabu Search, Genetic Algorithm, and Ant Colony Search. We compare against general metaheuristics because all previous algorithms for task sequencing are designed for a different objective function, time-to-threshold, and cannot be applied to cumulative return maximization. 


\subsection{Metaheuristic Adaptation}
\label{sec:meta}
Despite the generality of metaheuristc algorithms, all the one we chose must be adapted to the particular problem at hand. In this section we describe how each one has been tailored to the problem of task sequencing.

In Tabu Search (TS) \cite{glover1989tabu}, we create the neighborhood of a current curriculum by: generating a list of curricula $R$ composed of all the curricula obtained by removing from, or adding to, the last task in the current best curriculum; and generating all curricula resulting from any pairwise swap of any two tasks of a curriculum in $R$. We empty the tabu list of size $T$, when full, following a FIFO strategy. 
In our experiments $T = 30$.
 
For Genetic Algorithm (GA) \cite{goldberg1989genetic}, we set the initial population as $Q$ randomly sampled curricula from $\mathcal{C}_{\leq L}$. At each iteration we select two parent curricula with a roulette wheel selection.
Given two parent curricula we generate a new population of $Q$ candidate curricula by applying a standard single point cross over at randomized lengths along each parent gene (sequence of intermediate tasks). Each cross over step produces two children curricula and the process is repeated until $Q$ children curricula are created.
We also included a form of elitism in order to improve the performance of the algorithm by adding the parents to the population they generated. Genetic algorithms also include the definition of a mutation operator. In our implementation this acts on each candidate curriculum in the newly generated population with probability $p_m$. The mutation can be of two equally probable types: task-wise mutation, which given a candidate curriculum of length $l$, changes each of its intermediate tasks with probability equal to $1/l$; length-wise mutation, where equal probability is given to either dropping or adding a new source at a randomly selected position of a candidate curriculum. 
In our experiments $Q = 50$ and $p_m = 0.5$.

For Ant Colony Optimization (ACO) \cite{dorigo1991ant}, each agent in the colony moves towards the goal by adding a new intermediate task to the current candidate curriculum $c$ which represents the trail walked by the ant. Given a source task $m_i$ its probability of being selected is $P(m_i) = [(\tau_{m_i}+K)^\alpha + I_{m_i}^\beta] / [\sum_{E}{[(\tau_{m_j}+K)^\alpha + I_{m_j}^\beta}]]$ with $E = \{ m_j \in \mathcal{T} \vert m_j \notin c \}$. The variable $\tau_{m_i}$ represents the quantity of pheromone on task $m_i$ while following the current candidate curriculum $c$. The visibility $I_{m_i}$ is calculated as the performance improvement obtained by adding task $m_i$ to the current candidate curriculum when positive, and zero otherwise. Parameters $\alpha$ and $\beta$ control the influence of the pheromone versus the improvement, while $K$ is a threshold to control from what pheromone value the search starts to take it into account. The pheromone evaporation rate is specified with the parameter $\rho$ while the maximum level of pheromone to be accumulated over a candidate solution is set to $f_{max}$.  
In our experiments $\alpha = 1, \beta = 1.2, K = 5, f_{max} = 50, \rho = 0.2$ and the number of ants is $20$.
The parameters of all algorithms have been fine-tuned manually across all experiments.

\subsection{Domains}

We use two domains implemented within the software library Burlap~\footnote{http://burlap.cs.brown.edu}. Both domains have been previously used for CL \cite{leonetti17,narvekarIJCAI17,daSilva2018,foglino2019optimization}. 

\subsubsection{GridWorld}

GridWorld is an implementation of an episodic grid-world domain. Each cell can be free, or occupied by a  \emph{fire}, \emph{pit}, or \emph{treasure}. 
The agent can move in the four cardinal directions, and the actions are deterministic. The reward is $-2500$ for entering a pit, $-500$ for entering a fire, $-250$ for entering the cell next to a fire, and $200$ for entering a cell with the treasure. The reward is $-1$ in all other cases. The episodes terminate under one of these three
conditions: the agent falls into a pit, reaches the treasure, or executes a maximum number of actions. 

\subsubsection{BlockDude}

BlockDude is a puzzle game (Figure \ref{Fig:sample})  where the agent has to stack boxes in order to climb over walls and reach the exit. 
The available actions are moving left, right, up, pick up a box and put down a box. The agent receives a reward of $-1$ for each action taken, and an episode terminates when a maximum number of actions is executed. 

\subsection{Experiments}

For both domains we computed and analyzed all curricula within the given maximum length, so that, for each experiment, we know the globally optimal curriculum.

\begin{figure}[htbp]
	\includegraphics[width=\linewidth]{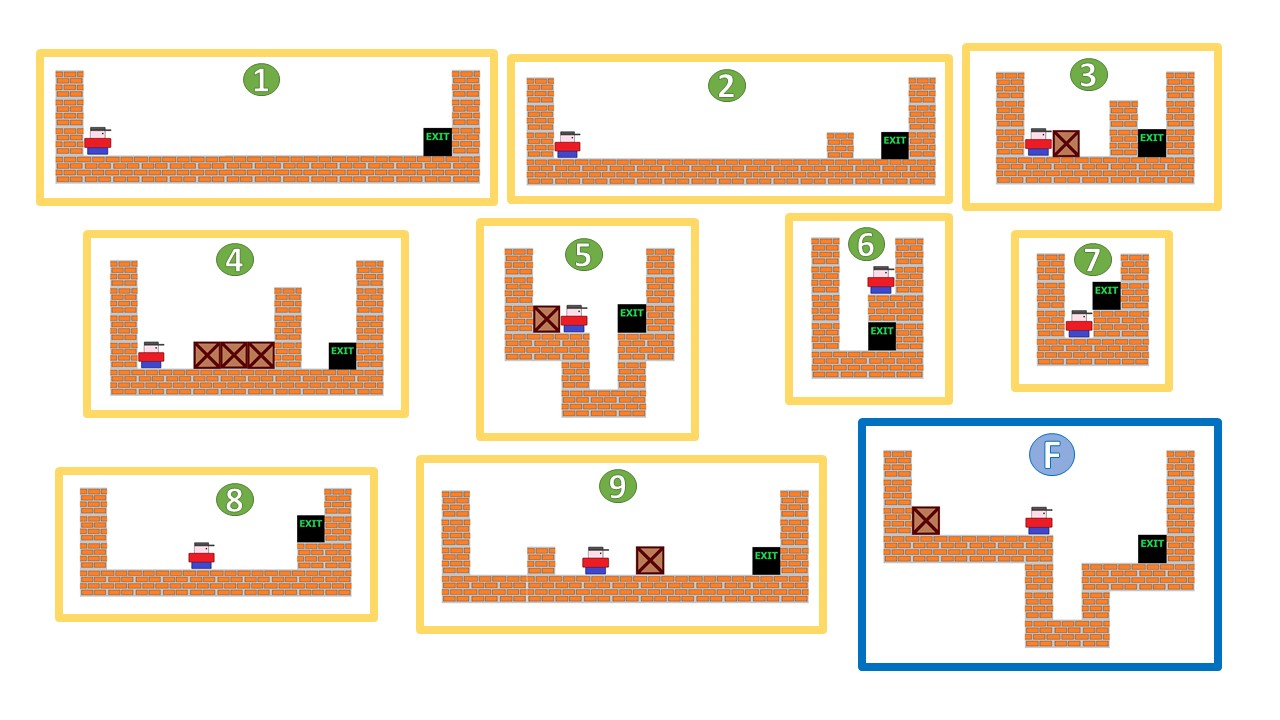}
      \caption{Intermediate (in yellow) and final task (in blue) of the second experiment in the BlockDude domain. The global optimum is the curriculum $7$-$1$-$5$.}
	\label{Fig:sample}
\end{figure}

\begin{figure*}[htbp]
	\centering
	\resizebox{\linewidth}{!}{
		\subfloat{\includegraphics{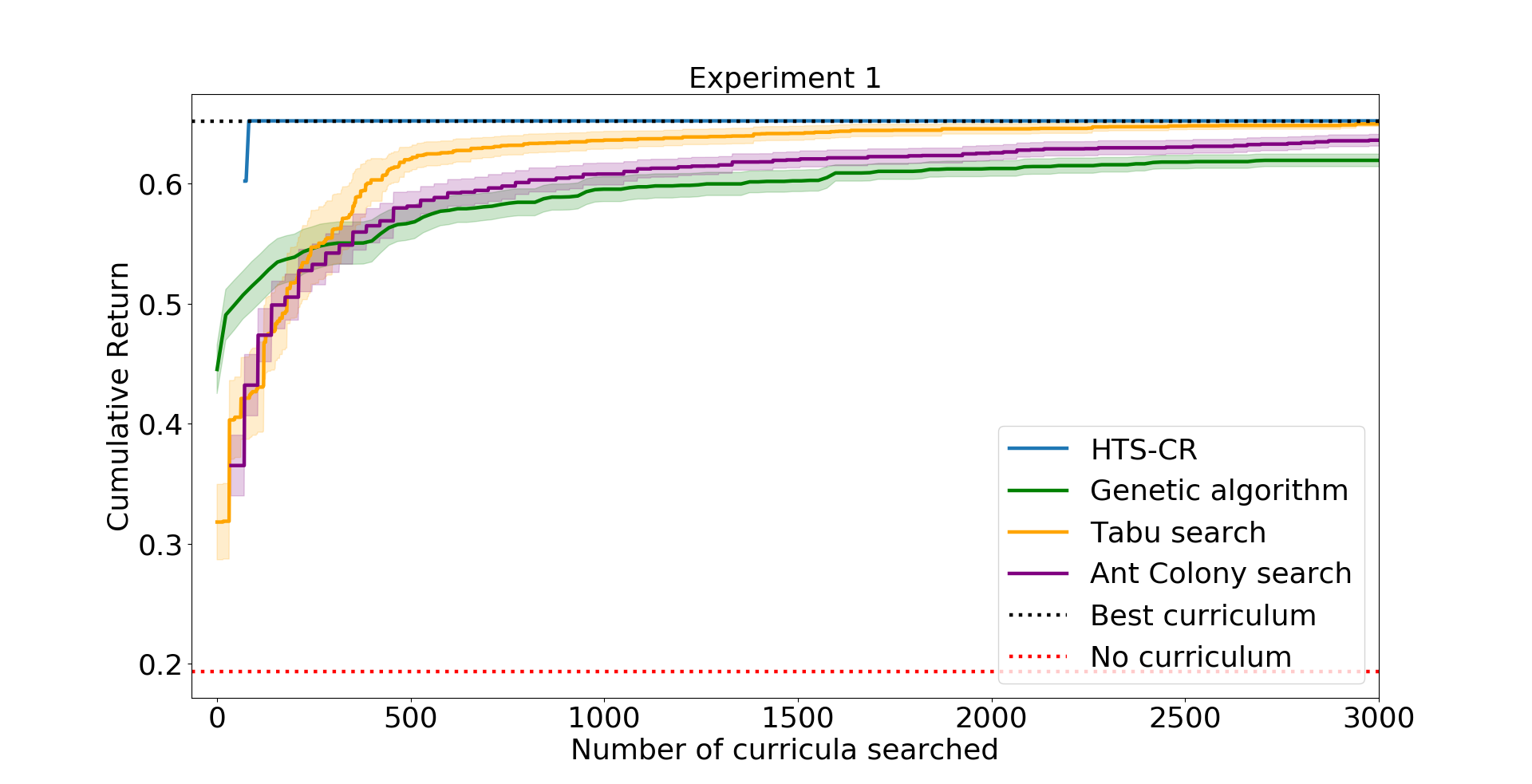}}\hfill
		\subfloat{\includegraphics{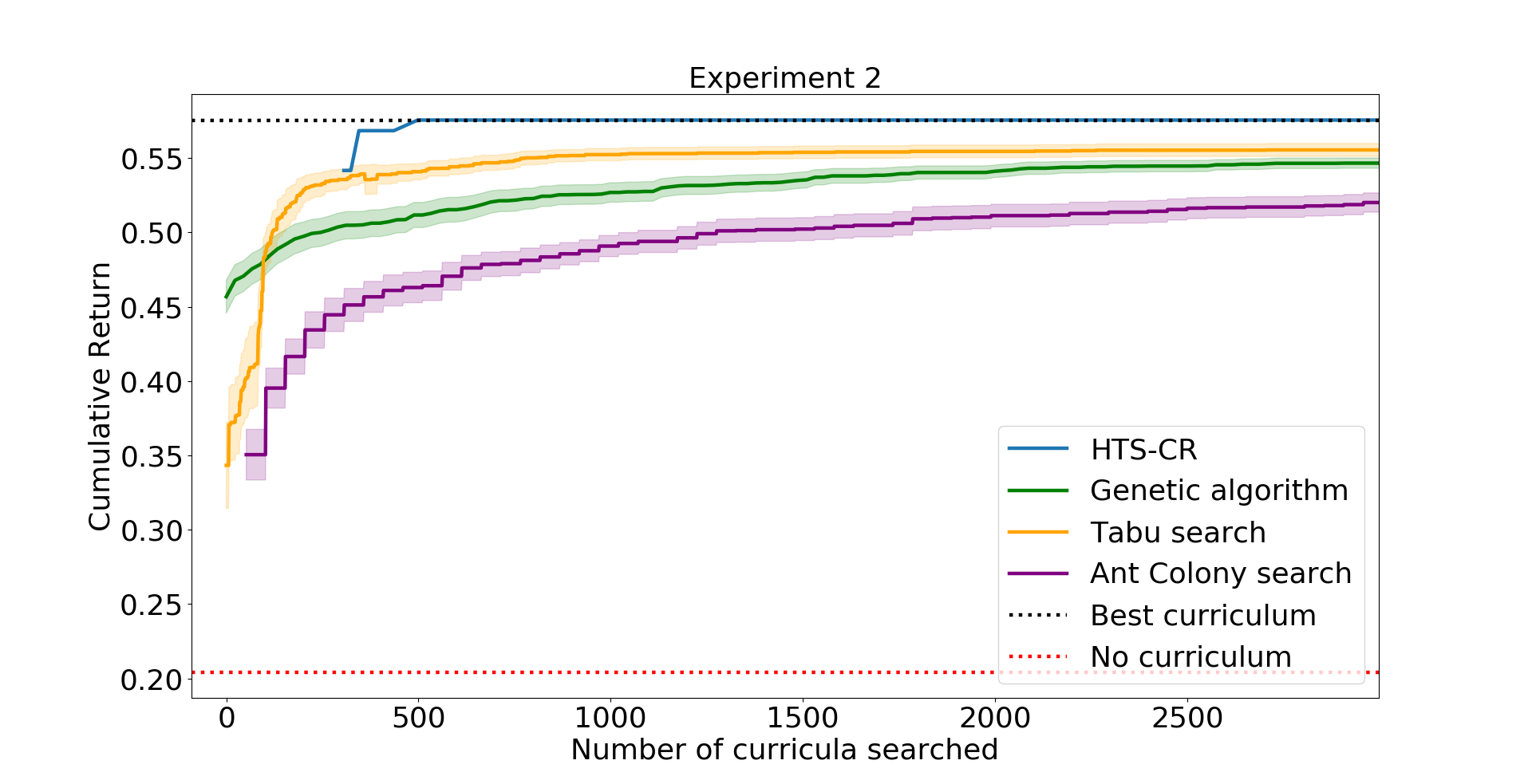}}
	}	
	\resizebox{\linewidth}{!}{
		\subfloat{\includegraphics{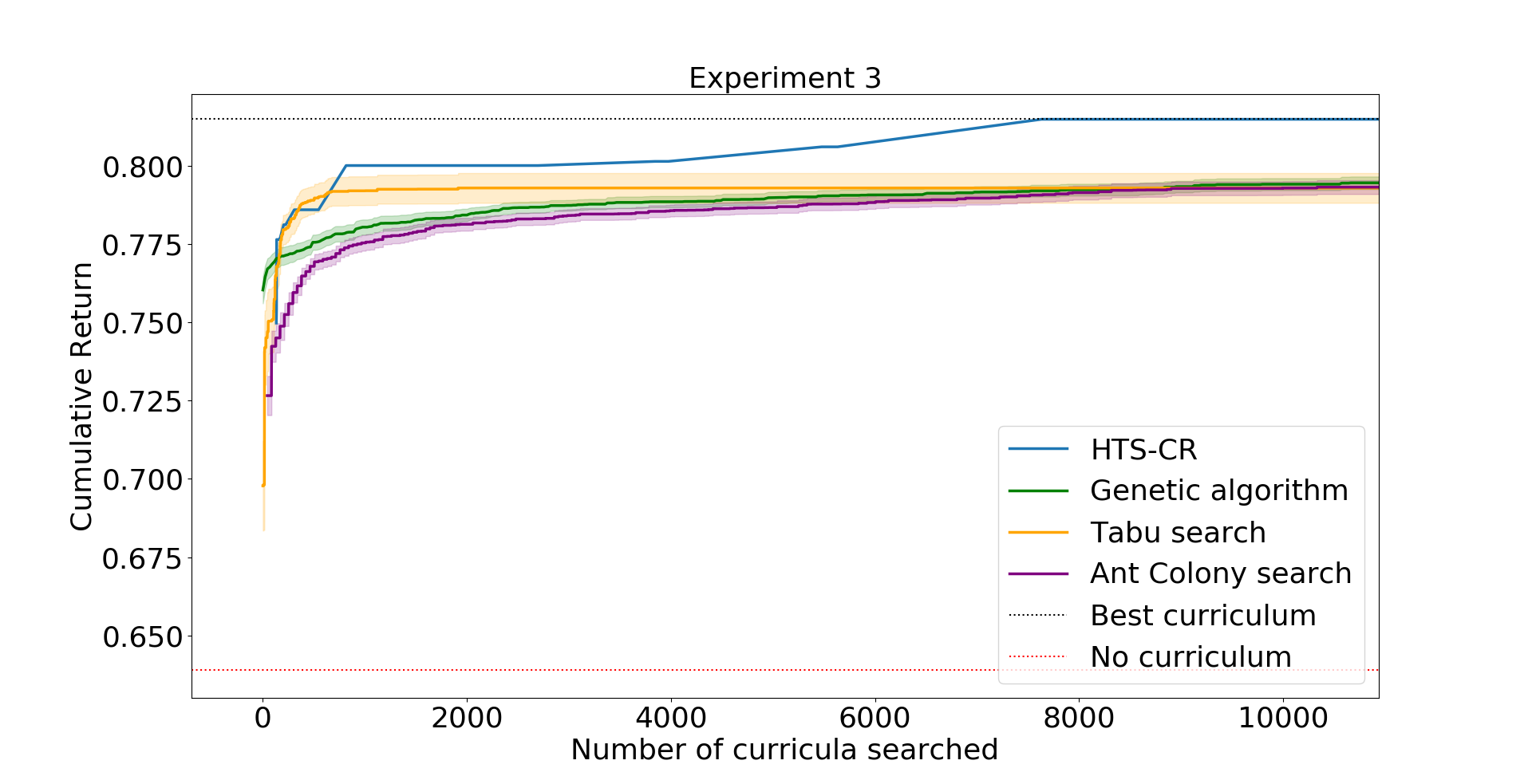}}\hfill
		\subfloat{\includegraphics{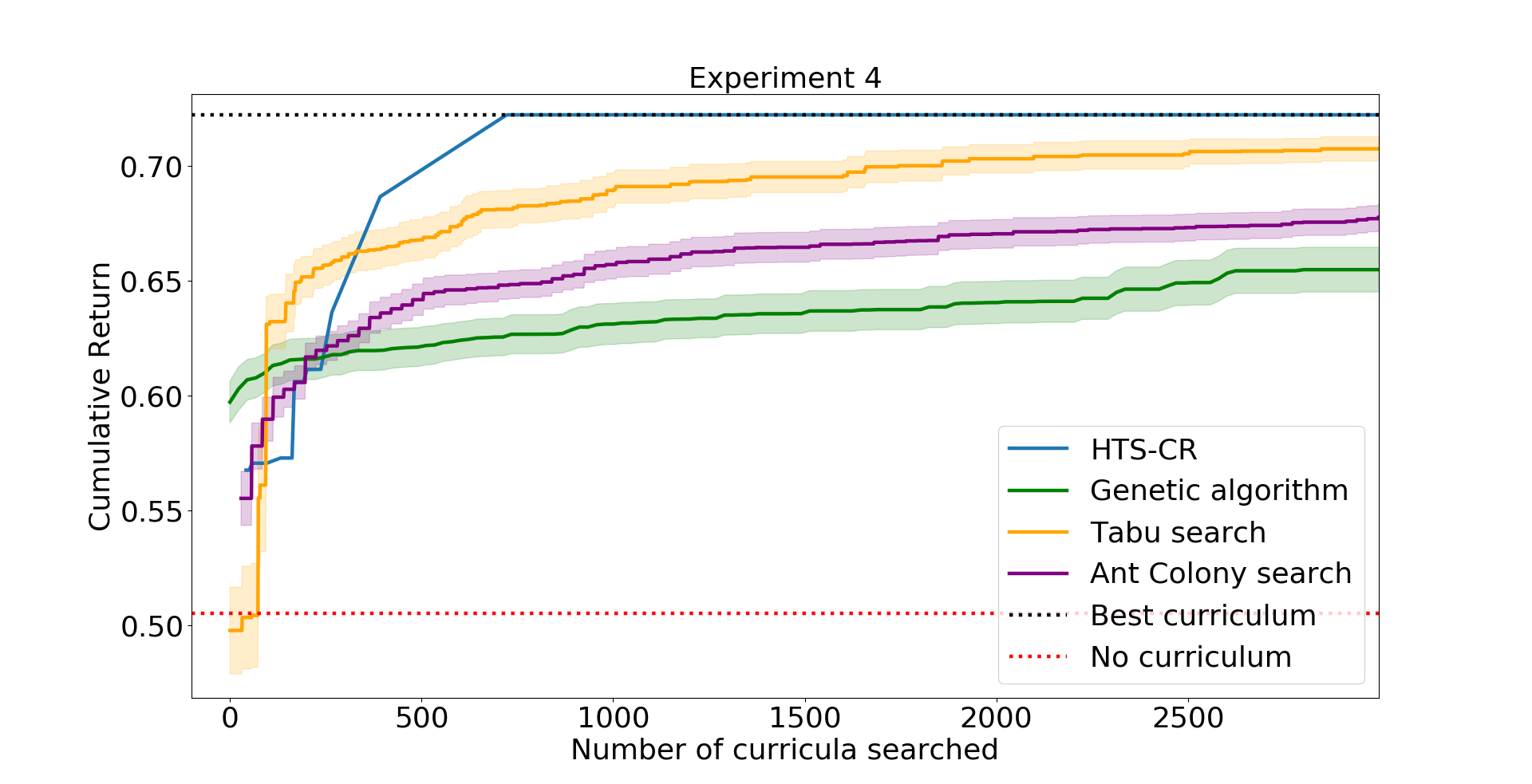}}
	}
	\caption{The results of the four experiments on the BlockDude (top) and Gridworld (bottom) domains. HTS-CR quickly outperforms the other algorithms finding the globally optimal curriculum.} \label{Fig:ExpsOpt}
\end{figure*}

We ran two sets of experiments per domain, one in which the number of tasks is high and the maximum length is low, and one in which, on the contrary, the number of tasks is low, but the maximum length is high. For BlockDude, Experiment $1$ (Figure \ref{Fig:ExpsOpt}) has parameters $n=18$ and $L=3$, while in Experiment $2$ $n=9$ and $L=5$. For GridWorld, Experiment $3$ (Figure \ref{Fig:ExpsOpt}) has parameters $n \coloneqq \lvert \mathcal{T} \rvert =12$ and $L=4$, while in Experiment $4$ $n=7$, and $L=7$. For both domains, the intermediate tasks have been generated manually using methods from \citeauthor{narvekarIJCAI17}~\shortcite{narvekarIJCAI17}, by varying the size of the environment, and adding and removing elements (pits and fires in GridWorld, and columns and movable blocks in BlockDude). We intentionally created both tasks that provide positive and negative transfer towards the final task, in order to test the ability of the sequencing algorithm to choose the most appropriate ones. In Figure \ref{Fig:sample} we show, as an example, the intermediate tasks and relative final task for the second experiment in the BlockDude domain. Each of the four experiments has a different final task and set of intermediate tasks.
All tasks are run for a number of episodes that ensures that the agent has converged to the optimal policy, determined at the time of task generation. Each curriculum  was repeated for $10$ epochs, and expected cumulative return approximated with the average. The agent learns with Sarsa($\lambda$) using Tile Coding, while exploring with $\epsilon$-greedy \footnote{The complete source code of all our experiments can be found at https://github.com/francescofoglino/Curriculum-Learning}.

In Figure \ref{Fig:ExpsOpt} we compare HTS-CR against the metaheuristic methods described in Section \ref{sec:meta}. The four plots, one for each experiment, show the value of the best curriculum over the number of curricula evaluated by each algorithm. Curricula were evaluated by having the agent learn each one multiple times, and averaging the results to estimate the objective in Equation \ref{eq:objective}. The cumulative return was normalized in $[0,1]$ for ease of comparison across the experiments, where $1$ is the return accumulated by the optimal policy at every episode.  As Tabu Search, Genetic Algorithm and Ant Colony Optimization are stochastic methods, their performance were averaged over $70$ runs and plotted showing the $95\%$ confidence interval. In all the experiments HTS-CR has an initial offset of $n(n-1)$ evaluations spent to consider all the possible pairs, whereas all the metaheuristics immediately start finding possible solutions. Nevertheless, HTS-CR quickly outperformed all the other algorithms, and always found the globally optimal curriculum the fastest, showing the benefit of the initial pair evaluations. Furthermore, our approach is deterministic and has no parameters.

\section{Real-world Validation}
The second experiment is aimed at validating the proposed use of curriculum learning in a realistic scenario. We set out to show that an appropriate curriculum can ``prepare'' the agent better than the traditional approach of learning the optimal policy in the simulated task and then transfer it.



\subsection{MGEnv Domain}
\label{sec:RealWorld}
MGEnv is a simulated micro-grid domain modeled out of real data from the PecanStreet Inc. database. The domain includes historical data about hourly energy consumption and solar energy generation of different buildings from January 2016 to December 2018. A task in this domain is defined by the combination of three elements: the model of the electrical device to optimize; the user's monthly schedule, specifying the days in which the user wishes to run the device; the month of simulation, with the energy generation and consumption of the given building. The device we used behaves as a time-shifting load: once started it runs for several time steps and cannot be interrupted before the end of its routine. This is the most challenging type of device present in the database. The goal is to find the best time of day to run the given device, optimizing direct use of the generated energy, while respecting the user's device schedule. The agent receives a reward of $30$ when the device energy consumption is fully covered by the energy generated by the building, $-10$ when energy not generated by the building is used and $-200$ if the device is  not run accordingly to the user schedule.


\subsection{Experiments}

We developed a DeepRL agent for the MGEnv domain with an Actor-Critic architecture, using Proximal Policy Optimization~\cite{schulman2017proximal}, and Progressive Neural Networks~\cite{rusu2016progressive}, for value function transfer.

All the tasks in this domain are for the control of the same electric device. We created $10$ final tasks using the same month and user schedule, but different buildings. We divided them into a training and test set of $5$ tasks each. Without CL, a natural approach would be to learn the optimal policy for one of the source tasks in the training set, and transfer the optimal value function to a target task in the test set.

We created $n = 5$ intermediate tasks, by selecting a combination of  schedule, month and building from a set of $3$ schedules, $5$ months and $3$ buildings. Some of these tasks are easier to learn than others, providing the basis for the curriculum.
We optimized the curriculum with HTS-CR over the training set with a maximum length $L = 4$, repeating each evaluation $5$ times to estimate the cumulative return. The best curriculum was found after $37$ curriculum evaluations. We then took the value function after the tail of the curriculum, before the final tasks, and transferred it to each of the $5$ tasks in the test set. We evaluated the performance of the agent initialized from the curriculum, and compared it to the behavior of the agent initialized from single-target transfer. 

\begin{figure}[h!]
	\includegraphics[width=\linewidth]{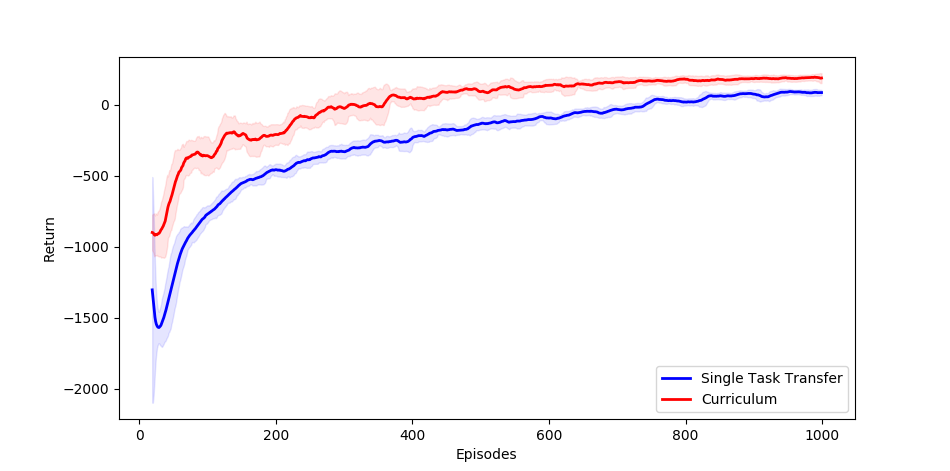}
	\caption{Average return over the test tasks in the MGEnv domain}
	\label{Fig:electPerf}
\end{figure}

Figure \ref{Fig:electPerf} shows the results of this experiment. The curriculum was generated using all the training tasks together, and evaluated over each one of the $5$ test tasks separately. Therefore, the plotted results are the average over $5$ runs with $95\%$ confidence intervals. Single-task transfer, on the other hand, was trained on each one of the training tasks and evaluated on each test task, resulting in the average over $25$ runs.  We show that, on average, initializing the learning from the curriculum achieves a higher cumulative return than initializing from the optimal value function of one of the source tasks. Our approach used $54\%$ less energy, on average, from external sources. The results show that a curriculum generated from simulated tasks can indeed generalize to similar, real, tasks (recall that all the data in these simulations is from real buildings) and provide a significant improvement of the behavior during learning.

\section{Conclusions}

We introduced a novel setting for curriculum learning in reinforcement learning, with the aim of shaping exploration in critical tasks. Furthermore, we introduced HTS-CR, a heuristic algorithm for task sequencing, and experimentally demonstrated that it outperforms several popular metaheuristc algorithms. We validated our approach on a home micro-grid controller, based on real data, showing that the knowledge provided by the curriculum is on average a more appropriate initialization than the optimal policy of a simulated task.

\section*{Acknowledgments}
This work has taken place in the Sensible Robots Research Group at the University of Leeds, which is partially supported by EPSRC (EP/R031193/1, EP/S005056/1).


\bibliographystyle{named}
\bibliography{bibfile}

\begin{thebibliography}{}

\bibitem[\protect\citeauthoryear{Asada \bgroup \em et al.\egroup
  }{1996}]{asada1996purposive}
Minoru Asada, Shoichi Noda, Sukoya Tawaratsumida, and Koh Hosoda.
\newblock Purposive behavior acquisition for a real robot by vision-based
  reinforcement learning.
\newblock {\em Machine learning}, 23(2-3):279--303, 1996.

\bibitem[\protect\citeauthoryear{Baranes and Oudeyer}{2013}]{baranes2013active}
Adrien Baranes and Pierre-Yves Oudeyer.
\newblock Active learning of inverse models with intrinsically motivated goal
  exploration in robots.
\newblock {\em Robotics and Autonomous Systems}, 61(1):49--73, 2013.

\bibitem[\protect\citeauthoryear{{Da Silva} and Costa}{2018}]{daSilva2018}
Felipe~Leno {Da Silva} and Anna Helena~Reali Costa.
\newblock Object-oriented curriculum generation for reinforcement learning.
\newblock In {\em International Conference on Autonomous Agents and Multiagent
  Systems (AAMAS)}, 2018.

\bibitem[\protect\citeauthoryear{Da~Silva and Costa}{2019}]{da2019survey}
Felipe~Leno Da~Silva and Anna Helena~Reali Costa.
\newblock A survey on transfer learning for multiagent reinforcement learning
  systems.
\newblock {\em Journal of Artificial Intelligence Research}, 64:645--703, 2019.

\bibitem[\protect\citeauthoryear{Dorigo \bgroup \em et al.\egroup
  }{1991}]{dorigo1991ant}
Marco Dorigo, Vittorio Maniezzo, and Alberto Colorni.
\newblock The ant system: An autocatalytic optimizing process, 1991.

\bibitem[\protect\citeauthoryear{Florensa \bgroup \em et al.\egroup
  }{2018}]{florensa2018goalgen}
Carlos Florensa, David Held, Xinyang Geng, and Pieter Abbeel.
\newblock Automatic goal generation for reinforcement learning agents.
\newblock In {\em International Conference on Machine Learning (ICML)}, 2018.

\bibitem[\protect\citeauthoryear{Foglino \bgroup \em et al.\egroup
  }{2019}]{foglino2019optimization}
Francesco Foglino, Christiano Coletto~Christakou, and Matteo Leonetti.
\newblock An optimization framework for task sequencing in curriculum learning.
\newblock In {\em Proceedings of 9th Joint IEEE International Conference on
  Development and Learning and on Epigenetic Robotics (ICDL-EpiRob)}, 2019.

\bibitem[\protect\citeauthoryear{Glover}{1989}]{glover1989tabu}
Fred Glover.
\newblock Tabu search—part i.
\newblock {\em ORSA Journal on computing}, 1(3):190--206, 1989.

\bibitem[\protect\citeauthoryear{Goldberg}{1989}]{goldberg1989genetic}
David~E Goldberg.
\newblock Genetic algorithms in search.
\newblock {\em Optimization, and MachineLearning}, 1989.

\bibitem[\protect\citeauthoryear{Jaksch \bgroup \em et al.\egroup
  }{2010}]{jaksch2010near}
Thomas Jaksch, Ronald Ortner, and Peter Auer.
\newblock Near-optimal regret bounds for reinforcement learning.
\newblock {\em Journal of Machine Learning Research}, 11(Apr):1563--1600, 2010.

\bibitem[\protect\citeauthoryear{Lazaric}{2012}]{lazaric2012transfer}
Alessandro Lazaric.
\newblock Transfer in reinforcement learning: a framework and a survey.
\newblock In {\em Reinforcement Learning}, pages 143--173. Springer, 2012.

\bibitem[\protect\citeauthoryear{Narvekar \bgroup \em et al.\egroup
  }{2016}]{leonetti16}
Sanmit Narvekar, Jivko Sinapov, Matteo Leonetti, and Peter Stone.
\newblock Source task creation for curriculum learning.
\newblock In {\em International Conference on Autonomous Agents and Multiagent
  Systems (AAMAS)}, 2016.

\bibitem[\protect\citeauthoryear{Narvekar \bgroup \em et al.\egroup
  }{2017}]{narvekarIJCAI17}
Sanmit Narvekar, Jivko Sinapov, and Peter Stone.
\newblock Autonomous task sequencing for customized curriculum design in
  reinforcement learning.
\newblock In {\em (IJCAI), The 2017 International Joint Conference on
  Artificial Intelligence}, 2017.

\bibitem[\protect\citeauthoryear{Riedmiller \bgroup \em et al.\egroup
  }{2018}]{pmlr-v80-riedmiller18a}
Martin Riedmiller, Roland Hafner, Thomas Lampe, Michael Neunert, Jonas Degrave,
  Tom van~de Wiele, Vlad Mnih, Nicolas Heess, and Jost~Tobias Springenberg.
\newblock Learning by playing solving sparse reward tasks from scratch.
\newblock In {\em Proceedings of the 35th International Conference on Machine
  Learning}, pages 4344--4353, 2018.

\bibitem[\protect\citeauthoryear{Rusu \bgroup \em et al.\egroup
  }{2016}]{rusu2016progressive}
Andrei~A Rusu, Neil~C Rabinowitz, Guillaume Desjardins, Hubert Soyer, James
  Kirkpatrick, Koray Kavukcuoglu, Razvan Pascanu, and Raia Hadsell.
\newblock Progressive neural networks.
\newblock {\em arXiv preprint arXiv:1606.04671}, 2016.

\bibitem[\protect\citeauthoryear{Ruvolo and Eaton}{2013}]{ruvolo2013ella}
Paul Ruvolo and Eric Eaton.
\newblock Ella: An efficient lifelong learning algorithm.
\newblock In {\em In Proceedings of the 30th International Conference on
  Machine Learning}, pages 507--515, 2013.

\bibitem[\protect\citeauthoryear{Schulman \bgroup \em et al.\egroup
  }{2017}]{schulman2017proximal}
John Schulman, Filip Wolski, Prafulla Dhariwal, Alec Radford, and Oleg Klimov.
\newblock Proximal policy optimization algorithms.
\newblock {\em arXiv preprint arXiv:1707.06347}, 2017.

\bibitem[\protect\citeauthoryear{Shao \bgroup \em et al.\egroup
  }{2018}]{shao2018micromancl}
K.~Shao, Y.~Zhu, and D.~Zhao.
\newblock Starcraft micromanagement with reinforcement learning and curriculum
  transfer learning.
\newblock {\em IEEE Transactions on Emerging Topics in Computational
  Intelligence}, pages 1--12, 2018.

\bibitem[\protect\citeauthoryear{Sukhbaatar \bgroup \em et al.\egroup
  }{2018}]{sukhbaatar2018}
Sainbayar Sukhbaatar, Zeming Lin, Ilya Kostrikov, Gabriel Synnaeve, Arthur
  Szlam, and Rob Fergus.
\newblock Intrinsic motivation and automatic curricula via asymmetric
  self-play.
\newblock In {\em International Conference on Learning Representations (ICLR)},
  2018.

\bibitem[\protect\citeauthoryear{Svetlik \bgroup \em et al.\egroup
  }{2017}]{leonetti17}
M~Svetlik, M~Leonetti, J~Sinapov, R~Shah, N~Walker, and P~Stone.
\newblock Automatic curriculum graph generation for reinforcement learning
  agents.
\newblock In {\em Thirty-First AAAI Conference on Artificial Intelligence}.
  Association for the Advancement of Artificial Intelligence, 2017.

\bibitem[\protect\citeauthoryear{Taylor and Stone}{2009}]{tl4rl}
Matthew~E Taylor and Peter Stone.
\newblock Transfer learning for reinforcement learning domains: A survey.
\newblock {\em Journal of Machine Learning Research}, 10(Jul):1633--1685, 2009.

\bibitem[\protect\citeauthoryear{Wilson \bgroup \em et al.\egroup
  }{2007}]{wilson2007multi}
Aaron Wilson, Alan Fern, Soumya Ray, and Prasad Tadepalli.
\newblock Multi-task reinforcement learning: a hierarchical bayesian approach.
\newblock In {\em Proceedings of the 24th international conference on Machine
  learning}, pages 1015--1022. ACM, 2007.

\bibitem[\protect\citeauthoryear{Wu and Tian}{2017}]{doom2017}
Yuxin Wu and Yuandong Tian.
\newblock Training agent for first-person shooter game with actor-critic
  curriculum learning.
\newblock In {\em International Conference on Learning Representations (ICLR)},
  2017.

\bibitem[\protect\citeauthoryear{Wu \bgroup \em et al.\egroup }{2018}]{wu2018}
Yuechen Wu, Wei Zhang, and Ke~Song.
\newblock Master-slave curriculum design for reinforcement learning.
\newblock In {\em Proceedings of the Twenty-Seventh International Joint
  Conference on Artificial Intelligence, {IJCAI-18}}, pages 1523--1529.
  International Joint Conferences on Artificial Intelligence Organization, 7
  2018.

\end{thebibliography}

\end{document}